\documentclass[times,review,10pt,numbers]{elsarticle}
\usepackage[
    a4paper,
    top=4.3cm,
    right=4.8cm,
    bottom=4.3cm,
    left=4.8cm
]{geometry}
\usepackage{hyperref}
\usepackage{amssymb}
\usepackage{amsmath}
\usepackage{algorithm}
\usepackage{algorithmic}
\usepackage{cleveref}

\usepackage[percent]{overpic}
\usepackage{booktabs}
\usepackage[table]{xcolor} 
\usepackage{multirow}
\definecolor{color_gray}{RGB}{229,229,229}
\definecolor{color_blue}{RGB}{252,182,165}
\definecolor{color_pink}{RGB}{255,217,178}
\definecolor{color_yellow}{RGB}{255,255,204}
\definecolor{color_blue1}{RGB}{135, 206, 235}
\usepackage{orcidlink}
\usepackage{color} 
\usepackage{caption}
\usepackage{placeins}
\usepackage{bm}
\usepackage{graphicx}
\AtBeginDocument{%
}

\begin{document}

\begin{frontmatter}
\title{%
\fontsize{14pt}{16.8pt}\selectfont
CFANet: Learning Conflict-Free Weather Adapters for All-Weather Image Restoration}

\author[a]{Minglong Xue} \ead{xueml@cqut.edu.cn}
\author[a]{Wenchao Yan} \ead{yanwenchao@stu.cqut.edu.cn}
\author[a]{Guoan Guan} \ead{gga@stu.cqut.edu.cn}
\author[b] {Palaiahnakote Shivakumara} \ead{S.Palaiahnakote@salford.ac.uk}

\author[c]{Yirui Wu\corref{cor1}}
\ead{wu.yirui@uaeu.ac.ae}

\affiliation[a]{
    organization={Chongqing University of Technology},
    city={Chongqing},
    postcode={400054},
    country={China}}
    
\affiliation[b]{
    organization={University of Salford},
    city={Manchester},
    postcode={M5 4WT},
    country={UK}}

\affiliation[c]{
    organization={United Arab Emirates University},
    city={Al Ain},
    state={Abu Dhabi},
    country={United Arab Emirates}}

\cortext[cor1]{Corresponding author}

\begin{abstract}
All-weather image restoration aims to recover clean images from diverse weather degradations such as rain, haze, and snow. Most existing all-in-one methods handle different weather types with a unified network or rely on single-label weather modeling. However, real-world adverse weather is often compositional, where multiple known weather primitives may coexist in a single image. To this end, we propose Conflict-Free Adapter Network (CFANet) for open-compositional all-weather image restoration. CFANet first employs a Weather Mixture Router (WMR) to predict non-exclusive composition weights for rain, haze, and snow from shallow features. Guided by these weights, a Shared-Private Weather Adapter (SPWA) dynamically composes a common adapter with weather-private adapters inside each Conflict-Free Adapter Block, preserving shared restoration capability while providing weather-specific residual adaptation. Moreover, a lightweight Natural Color Adapter (NCA) performs constrained color calibration to improve color fidelity and visual naturalness. Extensive experiments demonstrate that the proposed method achieves competitive adverse weather image restoration performance and benefits downstream visual tasks. Our code is available at {\url{https://github.com/wenchao-tech/CFANet-main}}.
\end{abstract}


\end{frontmatter}


\section{Introduction}
 
Adverse weather conditions, such as rain, haze, and snow, often degrade image visibility, distort color appearance, and corrupt local structures, which may further affect downstream vision systems~\cite{liu2022image, garcia2017review, mușat2021multi}. All-weather image restoration therefore aims to recover clean and faithful images from diverse weather-degraded inputs within a unified framework. Compared with task-specific restoration models, all-in-one methods are more flexible and practical because they avoid training a separate network for each weather type.

\begin{figure*}[t]
\centering
\centerline{\includegraphics[width=\textwidth,height=0.42\textwidth]{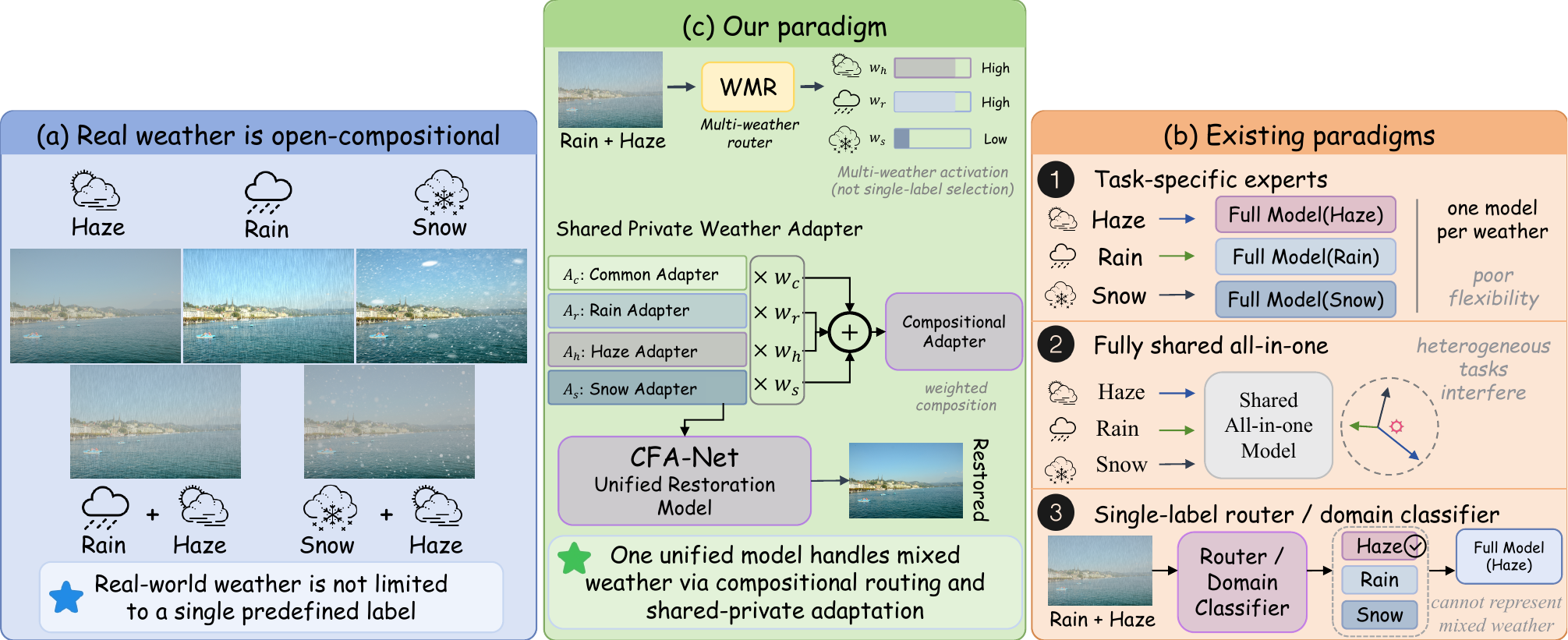}}
\caption{Motivation of CFANet. (a) Real-world adverse weather can be composed of multiple known weather primitives. (b) Existing paradigms either rely on task-specific experts, fully shared all-in-one models, or single-label routers, which are limited in flexibility, interference reduction, or compound-weather representation. (c) CFANet addresses this problem through non-exclusive weather
routing and shared-private adapter composition.}
\label{motivation}
\end{figure*}

However, real-world adverse weather is not always restricted to a single predefined category. As illustrated in Fig.~\ref{motivation}(a), multiple weather components may coexist in one image, such as rain with haze or snow with haze. This suggests that a degraded image can be better regarded as a flexible composition of known weather primitives rather than an instance of a mutually exclusive weather label. In this work, we refer to this setting as open-compositional all-weather restoration, where “open-compositional” denotes flexible combinations of known primitives, including rain, haze, and snow, instead of open-set recognition of unseen degradation categories.

Existing all-weather restoration paradigms remain limited under this compositional setting. As shown in Fig.~\ref{motivation}(b), task-specific expert models~\cite{zamir2022restormer, zamir2021multi} provide strong restoration ability for individual weather types but lack flexibility when multiple degradations appear simultaneously. Fully shared all-in-one models~\cite{valanarasu2022transweather} process all weather types with the same parameter space, which is efficient but may cause interference among heterogeneous restoration directions. Single-label routers or domain classifiers~\cite{potlapalli2023promptir,li2022all} further restrict the restoration process by selecting one dominant weather type, making them unsuitable for compound-weather degradations where multiple weather components should be activated together. These limitations motivate a composition-aware restoration framework that can activate multiple weather-specific directions while preserving shared restoration capability.

To this end, we propose Conflict-Free Adapter Network (CFANet), a unified restoration framework for open-compositional all-weather image restoration. As shown in Fig.~\ref{motivation}(c), CFANet first introduces a Weather Mixture Router (WMR) to predict non-exclusive composition weights for rain, haze, and snow from shallow image features. Unlike softmax-based single-label routing, WMR adopts non-exclusive weather activation, allowing multiple weather primitives to be active simultaneously. Guided by these weights, CFANet embeds a Shared-Private Weather Adapter (SPWA) into each Conflict-Free Adapter Block, where a common adapter captures weather-agnostic restoration patterns and weather-private adapters model weather-specific residual directions. Through dynamic adapter composition, CFANet can adjust its restoration behavior according to flexible weather combinations within a single unified model.

In addition, we introduce a lightweight Natural Color Adapter (NCA) to alleviate color shifts that commonly appear in all-weather restoration. NCA performs constrained color calibration on the restored image, improving color fidelity and visual naturalness without introducing an additional heavy restoration branch. We further formulate a conflict-free learning objective to supervise weather composition prediction and encourage private adapters to learn distinguishable restoration directions. The main contributions of this work are summarized as follows:
\begin{itemize}
\item We propose CFANet for open-compositional all-weather image restoration, where WMR models flexible weather compositions through non-exclusive routing over known weather primitives.

\item We design a Shared-Private Weather Adapter (SPWA) within the Conflict-Free Adapter Block, which dynamically composes a common adapter and weather-private adapters to capture shared restoration capability and weather-specific residual directions.

\item We introduce a lightweight Natural Color Adapter (NCA) and a conflict-free learning objective to improve color fidelity, enhance visual naturalness, and encourage disentangled weather-specific adaptation.
\end{itemize}

\section{Related Work}
\subsection{Unified All-Weather Restoration}
Adverse weather image restoration aims to recover clean images from rainy, hazy, and snowy inputs. Early methods~\cite{chen2021all, he2010single} usually design task-specific models for deraining, dehazing, or desnowing. Although these methods can exploit degradation-specific priors and achieve promising performance on individual tasks, they require separate models for different weather types and lack flexibility in complex real-world scenarios.

To improve model generality, all-in-one and all-weather restoration methods~\cite{li2025all} have been proposed to handle multiple degradations within a unified framework. All-in-One Bad Weather Removal~\cite{li2020all} explores neural architecture search for unified adverse weather removal, while TransWeather~\cite{valanarasu2022transweather} adopts a Transformer-based encoder-decoder with weather queries to process different degradations. AirNet~\cite{li2022all} learns degradation-aware representations, and PromptIR~\cite{potlapalli2023promptir} introduces learnable prompts to adapt the restoration process. Recent methods, such as Histoformer~\cite{sun2024restoring} and TANet~\cite{wang2024tanet}, further enhance unified restoration through histogram self-attention and triplet attention. Language-guided methods~\cite{luo2024controlling,conde2024instructir} and AWRaCLe~\cite{rajagopalan2025awracle} also introduce textual priors or visual in-context examples to provide explicit degradation guidance.

Despite their progress, most existing methods mainly focus on predefined weather tasks and usually assume a single dominant degradation type. However, real-world adverse weather can be compositional, where multiple known weather primitives may coexist and require different restoration directions. In contrast, CFANet explicitly models flexible weather compositions through weather mixture routing and shared-private adapter composition, supporting both single-weather and compound-weather restoration in a unified model.

\subsection{Condition-Guided Image Restoration}

Condition-guided image restoration improves the adaptability of unified models by introducing degradation-related conditions into the restoration process. Prompt-based methods, such as PromptIR~\cite{potlapalli2023promptir}, learn degradation-aware prompts to modulate network features, while instruction- and language-guided methods use textual or vision-language priors to specify restoration goals. Representative methods such as InstructIR~\cite{conde2024instructir} and DA-CLIP~\cite{luo2024controlling} further exploit natural language instructions or CLIP-based degradation representations.  In adverse weather restoration, LDR~\cite{yang2024language} uses pretrained vision-language priors to guide degradation estimation and pixel-wise top-K expert selection, including restoration under mixed weather. AWRaCLe~\cite{rajagopalan2025awracle} provides explicit degradation guidance through visual in-context examples.

Although effective, these methods~\cite{wu2025freprompter,sun2024restoring,xie2026all} usually rely on external prompts, language instructions, context examples, or pretrained vision-language models. In contrast, CFANet directly predicts weather composition weights from shallow features of the degraded image, enabling composition-aware restoration without additional guidance inputs.

\subsection{Composite Degradation Restoration}
Composite degradation restoration aims to recover clean images when multiple degradations coexist within the same input. Existing approaches address this problem through scene-level conditioning, degradation representation, and expert composition. OneRestore~\cite{guo2024onerestore} introduces CDD-11 and guides restoration using scene descriptors supplied through text or obtained automatically from visual attributes. RAHC~\cite{wan2025learning} investigates hybrid adverse weather restoration through multiple degradation representation subspaces and output-space discrimination, and introduces the HAC dataset. At the expert level, CWE-Net~\cite{shehzad2026cwe} employs multi-label weather routing and cross-weather fusion to combine specialized restoration responses. 

CFANet incorporates weather composition into the feature transformations within restoration blocks: non-exclusive weights for rain, haze, and snow modulate weather-private adapter outputs alongside an always-active common adapter. This organization combines shared restoration knowledge with weather-specific residual adaptation throughout the backbone.

\subsection{Mixture-of-Experts and Shared-Private Adaptation}

Balancing shared representation and task-specific specialization is important for unified image restoration. Fully shared networks are parameter-efficient, but may suffer from negative transfer when different degradations require distinct restoration behaviors. 

Existing methods alleviate this problem through expert mechanisms, parameter partitioning, or task-aware modulation. Mixture-of-Experts methods~\cite{zhu2023learning,park2025towards} use gating networks to select or fuse expert branches, while shared-private learning decomposes parameters into common and task-specific components to reduce task interference~\cite{ma2018modeling}. In adverse weather restoration, WGWS-Net~\cite{zhu2023learning} adopts a two-stage training strategy that first learns weather-general features and then adaptively expands weather-specific parameters. Adapter-based methods~\cite{chen2022adaptformer} further provide lightweight task-specific modulation within unified models. 

Different from these methods, CFANet introduces a shared-private weather adapter mechanism with non-exclusive composition routing. It dynamically composes a common adapter and weather-private adapters according to the input weather composition, enabling composition-aware restoration while reducing interference among heterogeneous weather directions.

\section{Methodology}
\begin{figure*}[t]
\centering
\centerline{\includegraphics[width=\textwidth,height=0.58\textwidth]{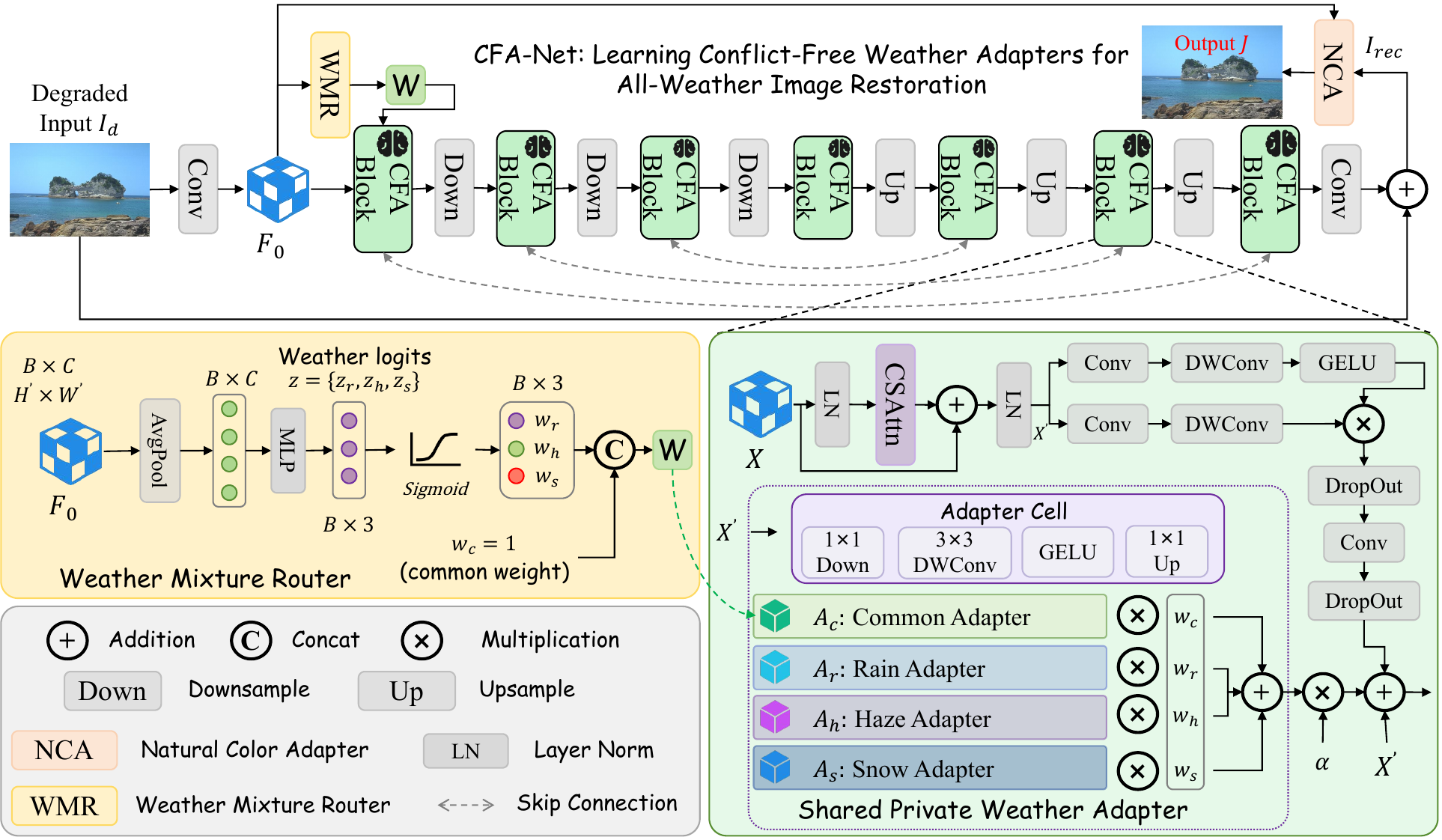}}
\caption{Overview of the proposed CFANet framework. The Weather Mixture Router (WMR) predicts image-level, non-exclusive rain, haze, and snow weights from shallow features $F_0$. These weights, together with a fixed common weight $w_c=1$, condition the Shared-Private Weather Adapter (SPWA) in each Conflict-Free Adapter Block (CFABlock) of the encoder--decoder backbone. SPWA combines common and weather-private adapter outputs for residual adaptation. The Natural Color Adapter (NCA) then calibrates the reconstructed image $I_{\mathrm{rec}}$ to obtain $I_{\mathrm{out}}$.}
\label{network}

\end{figure*}

\subsection{Problem Formulation}

Given a weather-degraded image $I_d$, all-weather image restoration aims to estimate a clean image $I_{out}$ that is close to the ground-truth image $I_{gt}$. We study open-compositional all-weather restoration, where \emph{open-compositional} refers to flexible combinations of known weather primitives rather than open-set recognition of unseen degradation categories. The weather primitive set is defined as
\begin{equation}
\mathcal{T}=\{\text{rain}, \text{haze}, \text{snow}\}.
\end{equation}

Different from single-label weather modeling, each degraded image is assigned a weak multi-hot composition label:
\begin{equation}
\mathbf{y}=[y_r,y_h,y_s]\in\{0,1\}^{3}
\end{equation}
where $y_r$, $y_h$, and $y_s$ indicate the presence of rain, haze, and snow, respectively. Single-weather images activate one entry, while compound-weather images can activate multiple entries. The model is therefore expected to adapt its restoration behavior according to the input weather composition while reducing interference among weather-specific restoration directions.

\subsection{Overall Framework}

As shown in Fig.~\ref{network}, CFANet is a hierarchical encoder-decoder restoration framework equipped with weather composition-aware adaptation. Given a degraded image $I_d$, CFANet first extracts shallow features:
\begin{equation}
F_0 = \mathrm{Conv}(I_d).
\end{equation}

The Weather Mixture Router (WMR) then predicts weather logits $\mathbf{z}$ and adapter composition weights $\mathbf{w}$ from $F_0$:
\begin{equation}
\mathbf{z}, \mathbf{w} = \mathrm{WMR}(F_0),
\end{equation}
where $\mathbf{z}$ represents weather responses and $\mathbf{w}$ controls the dynamic composition of shared and weather-private adapters.

The backbone contains multi-scale encoders, a bottleneck, decoders, and a refinement stage, all built with Conflict-Free Adapter Blocks (CFABlocks). The adapter weights $\mathbf{w}$ are shared across these blocks to inject composition-aware residual adaptation. After refinement, a reconstruction head predicts a residual image from the restored feature $F_{out}$:
\begin{equation}
I_{rec} = I_d + \mathrm{Conv}(F_{out}).
\end{equation}
Finally, the Natural Color Adapter (NCA) calibrates $I_{rec}$ with the shallow feature $F_0$ and produces the final output:
\begin{equation}
I_{out} = \mathrm{NCA}(I_{rec}, F_0).
\end{equation}

\subsection{Weather Mixture Router}

To model flexible weather compositions, CFANet introduces a Weather Mixture Router (WMR) after shallow feature extraction. Given the shallow feature $F_0$, WMR first applies global average pooling to obtain an image-level weather descriptor:
\begin{equation}
\mathbf{g} = \mathrm{GAP}(F_0).
\end{equation}

A lightweight MLP is then used to predict the logits of three known weather primitives:
\begin{equation}
\mathbf{z}=[z_r,z_h,z_s]=\mathrm{MLP}(\mathbf{g}),
\end{equation}
where $z_r$, $z_h$, and $z_s$ correspond to rain, haze, and snow, respectively.
Different from softmax-based routers that enforce mutually exclusive decisions, WMR adopts sigmoid activation to estimate non-exclusive weather weights:
\begin{equation}
[w_r,w_h,w_s]=\sigma(\mathbf{z}).
\end{equation}

For single-weather inputs, one private weight typically dominates, whereas compound-weather inputs can retain multiple active weights, allowing the corresponding private adapters to contribute jointly. To preserve weather-agnostic restoration capability, we further introduce a fixed common weight $w_c=1$ and concatenate it with the weather-private weights:
\begin{equation}
\mathbf{W}=[w_c,w_r,w_h,w_s].
\end{equation}

The resulting adapter weights $\mathbf{W}$ are shared across subsequent CFABlocks to guide the dynamic composition of common and weather-private adapters.

\subsection{Conflict-Free Adapter Block}

The Conflict-Free Adapter Block (CFABlock) is the basic restoration unit of CFANet. It preserves the generic feature transformation ability of the backbone while introducing weather composition-aware residual adaptation. Given an input feature $\mathbf{X}$ and the adapter weights $\mathbf{W}$ predicted by WMR, CFABlock first performs contextual modeling with layer normalization and compressed self-attention:
\begin{equation}
\mathbf{X}' = \mathbf{X} + \mathrm{CSAttn}(\mathrm{LN}(\mathbf{X})).
\end{equation}

The updated feature is then processed by a feed-forward network and the Shared-Private Weather Adapter (SPWA):
\begin{equation}
\mathbf{u} = \mathrm{LN}(\mathbf{X}'),
\end{equation}
\begin{equation}
\mathbf{X}'' =
\mathbf{X}' +
\mathrm{FFN}(\mathbf{u}) +
\alpha \cdot \mathrm{SPWA}(\mathbf{u}, \mathbf{W}),
\end{equation}
where $\alpha$ is a learnable scale that controls the contribution of the adapter branch.

SPWA consists of one common adapter and three weather-private adapters:
\begin{equation}
{A}=\{A_c,A_r,A_h,A_s\},
\end{equation}
where $A_c$ captures weather-agnostic restoration patterns, and $A_r$, $A_h$, and $A_s$ correspond to rain, haze, and snow residual adaptation, respectively. Each adapter adopts a lightweight bottleneck structure with a $1\times1$ down-projection, a $3\times3$ depth-wise convolution, GELU activation, and a $1\times1$ up-projection.

Given the adapter weights $\mathbf{W}=[w_c,w_r,w_h,w_s]$, SPWA dynamically composes the adapter outputs:
\begin{equation}
\begin{aligned}
\mathrm{SPWA}(\mathbf{u},\mathbf{W})
={}& w_c A_c(\mathbf{u})
+ w_r A_r(\mathbf{u}) \\
&+ w_h A_h(\mathbf{u})
+ w_s A_s(\mathbf{u}).
\end{aligned}
\label{eq:spwa}
\end{equation}

where $A_c$ serves as a shared restoration basis that captures weather-agnostic patterns, while $A_r$, $A_h$, and $A_s$ model weather-private residual directions for rain, haze, and snow, respectively. Guided by the composition weights from WMR, SPWA dynamically combines these adapters to support composition-aware feature restoration. This shared-private design helps reduce interference among heterogeneous weather degradations by separating common restoration capability from weather-specific residual adaptation.

\begin{figure}[t]
\centering
\centerline{\includegraphics[width=0.56\textwidth,height=0.4\textwidth]{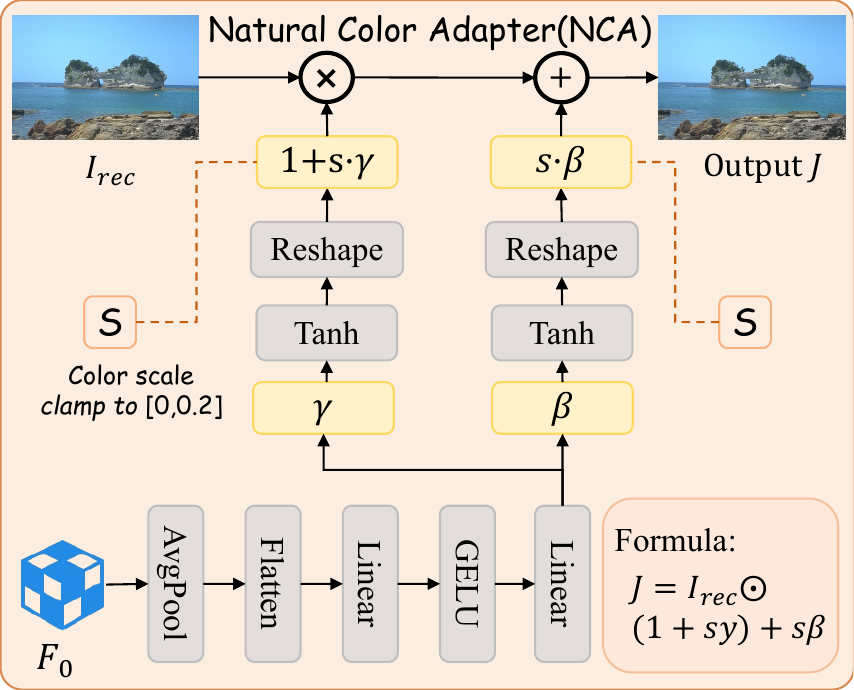}}
\caption{Architecture of the Natural Color Adapter (NCA). Global average pooling and an MLP map shallow features $F_0$ to channel-wise scale and bias parameters $\gamma$ and $\beta$, bounded by $\tanh$. NCA applies affine calibration to the reconstructed image as $J=I_{\mathrm{rec}}\odot(1+s\gamma)+s\beta$, where $s\in[0,0.2]$ controls the modulation magnitude and $\odot$ denotes element-wise multiplication.}
\label{NCA}

\end{figure}

\subsection{Natural Color Adapter}

Although CFABlocks perform composition-aware restoration, color shifts may still appear in all-weather restoration, especially under haze and snow degradations. To improve color fidelity and visual naturalness, we introduce a lightweight Natural Color Adapter (NCA) after the main restoration backbone.

Given the intermediate restored image $I_{rec}$ and the shallow feature $F_0$, NCA predicts channel-wise color modulation parameters from $F_0$. Specifically, $F_0$ is compressed by global average pooling and projected by two linear layers:
\begin{equation}
\mathbf{q}=\mathrm{Linear}_2(\mathrm{GELU}(\mathrm{Linear}_1(\mathrm{GAP}(F_0)))).
\end{equation}
The output $\mathbf{q}$ is split into two groups, $\gamma$ and $\beta$, which are reshaped to $3\times1\times1$ and constrained by $\tanh$. NCA then performs constrained affine color calibration:
\begin{equation}
I_{out}=I_{rec}\odot(1+s\gamma)+s\beta,
\end{equation}
where $\odot$ denotes element-wise multiplication, and $s$ is constrained within $[0,0.2]$ to avoid over-correction. In this way, NCA acts as a lightweight output calibration module, improving color consistency while preserving the main structural restoration result.

\subsection{Conflict-Free Learning Objective}

To stabilize the optimization of CFANet, we train the network with reconstruction, routing, and adapter disentanglement loss. For the restoration task, we adopt a Charbonnier-like reconstruction loss:
\begin{equation}
\mathcal{L}_{rec}
=
\frac{1}{N}\sum_{i=1}^{N}
\left(
\sqrt{(I_{out}^{i}-I_{gt}^{i})^2+c^2}-c
\right),
\end{equation}
where $I_{out}$ and $I_{gt}$ denote the restored and ground-truth images, respectively, and $c=0.03$.

To supervise the Weather Mixture Router, we use weak multi-hot weather composition labels:
\begin{equation}
\mathcal{L}_{router}
=
\mathrm{BCEWithLogits}(\mathbf{z},\mathbf{y}),
\end{equation}
where $\mathbf{z}=[z_r,z_h,z_s]$ and $\mathbf{y}=[y_r,y_h,y_s]$ correspond to rain, haze, and snow. We further impose a composition consistency loss between the predicted private adapter weights $\mathbf{w}_p=[w_r,w_h,w_s]$ and the weather label:
\begin{equation}
\mathcal{L}_{comp}
=
\left\|
\frac{\mathbf{w}_p}{\sum_{k=1}^{3} w_{p,k}+\epsilon}
-
\frac{\mathbf{y}}{\sum_{k=1}^{3} y_k+\epsilon}
\right\|_1 .
\end{equation}

To reduce redundancy among weather-private adapters, we apply an orthogonality constraint:
\begin{equation}
\mathcal{L}_{ortho}=
\frac{1}{K(K-1)}
\sum_{i\neq j}
\left|
\frac{\theta_i^{\top}\theta_j}
{|\theta_i|_2|\theta_j|_2+\epsilon}
\right|,
\end{equation}
where $\theta_i$ denotes the vectorized parameters of the $i$-th private adapter and $K=3$. 

The final training objective is
\begin{equation}
\mathcal{L}_{total}
=
\mathcal{L}_{rec}
+
\lambda_r \mathcal{L}_{router}
+
\lambda_c \mathcal{L}_{comp}
+
\lambda_o \mathcal{L}_{ortho},
\end{equation}
where $\lambda_r=0.01$, $\lambda_c=0.01$, and $\lambda_o=0.001$.

\begin{algorithm}[t]
\caption{One training iteration of CFANet}
\label{alg:cfanet}
\begin{algorithmic}[1]
\REQUIRE degraded image $I_d$, ground-truth image $I_{gt}$, weather label 
$\mathbf{y}=[y_r,y_h,y_s]$; shallow convolution $\mathrm{Conv}_{s}$; 
Weather Mixture Router $\mathrm{WMR}$; CFABlocks $\{\mathcal{B}_l\}$; 
Natural Color Adapter $\mathrm{NCA}$; loss weights $\lambda_r,\lambda_c,\lambda_o$
\STATE $F_0 \leftarrow \mathrm{Conv}_{s}(I_d)$
\STATE $\mathbf{g}\leftarrow \mathrm{GAP}(F_0)$;\;
       $\mathbf{z}=[z_r,z_h,z_s]\leftarrow \mathrm{MLP}(\mathbf{g})$
\STATE $[w_r,w_h,w_s]\leftarrow \sigma(\mathbf{z})$;\;
       $\mathbf{W}\leftarrow [w_c,w_r,w_h,w_s]$, where $w_c=1$
\STATE $X\leftarrow F_0$
\FOR{each CFABlock $\mathcal{B}_l$ in the encoder-decoder backbone}
    \STATE $X'\leftarrow X+\mathrm{CSAttn}(\mathrm{LN}(X))$
    \STATE $u\leftarrow \mathrm{LN}(X')$
    \STATE $A(u,\mathbf{W})\leftarrow 
    w_cA_c(u)+w_rA_r(u)+w_hA_h(u)+w_sA_s(u)$
    \STATE $X\leftarrow X'+\mathrm{FFN}(u)+\alpha A(u,\mathbf{W})$
\ENDFOR
\STATE $I_{rec}\leftarrow I_d+\mathrm{Conv}(X)$
\STATE $I_{out}
\leftarrow \mathrm{NCA}(I_{rec},F_0)$

\STATE Compute
$\mathcal{L}_{rec}$,
$\mathcal{L}_{router}$,
$\mathcal{L}_{comp}$, and
$\mathcal{L}_{ortho}$

\STATE $\mathcal{L}\leftarrow
\mathcal{L}_{rec}
+\lambda_r\mathcal{L}_{router}
+\lambda_c\mathcal{L}_{comp}
+\lambda_o\mathcal{L}_{ortho}$

\STATE update CFANet parameters by gradient descent on $\mathcal{L}$

\end{algorithmic}
\end{algorithm}

\subsection{Training Procedure}
Algorithm~\ref{alg:cfanet} summarizes one training iteration of CFANet. Given a degraded image, its clean target, and a multi-hot weather label, the network predicts weather composition weights, performs restoration through shared-private adapter composition, and applies output color calibration. The network is jointly optimized using the reconstruction, routing, composition consistency, and adapter orthogonality losses.

\section{Experiments}
\subsection{Experimental Settings}
\subsubsection{Datasets.}
For training, we use three representative paired datasets covering deraining, dehazing, and desnowing: Snow100K~\cite{liu2018desnownet} with 50,000 snowy-clean pairs, SRD~\cite{zamir2021multi} with 13,712 rainy-clean pairs, and OTS from RESIDE~\cite{li2018benchmarking} with 72,135 hazy-clean pairs, together with the training split of CDD-11~\cite{guo2024onerestore} for compound-weather restoration.

For synthetic evaluation, we use SOTS-Outdoor~\cite{li2018benchmarking}, Rain100H and Rain100L~\cite{yang2017deep}, and the Snow100K test set~\cite{liu2018desnownet} to evaluate single-weather restoration. To further assess compound-weather restoration, we include CDD-11, which contains multiple composite degradation types such as {\textit{haze+rain}} and {\textit{haze+snow}}. For real-world evaluation without clean references, we use RTTS~\cite{li2018benchmarking} and Fattal~\cite{fattal2014dehazing} for dehazing, SPA-DATA~\cite{wang2019spatial} for deraining, and RealSnow10K~\cite{lai2025snowmaster} for desnowing.

\subsubsection{Compared Methods and Evaluation Metrics.}
We compare CFANet with representative all-weather restoration methods, including WeatherDiff~\cite{ozdenizci2023restoring}, PromptIR~\cite{potlapalli2023promptir}, DA-CLIP~\cite{luo2024controlling}, GridFormer~\cite{wang2024gridformer}, DiffUIR~\cite{zheng2024selective}, OneRestore~\cite{guo2024onerestore}, OKNet~\cite{cui2024omni}, BioIR~\cite{cui2026bio}, CycleRDM~\cite{xue2025unified}, UniRestore~\cite{chen2025unirestore}, AWRaCLe~\cite{rajagopalan2025awracle}, AdaIR~\cite{cui2025adair}, StarIR~\cite{cui2026starir}, and WeatherRemover~\cite{qu2025weatherremover}. Since this work focuses on unified all-weather restoration, task-specific methods are not included in the main comparison.

For synthetic paired datasets, we report PSNR and SSIM~\cite{wang2004image} for fidelity and structural consistency, together with LPIPS~\cite{zhang2018unreasonable} and FID~\cite{heusel2017gans} for perceptual similarity and distribution consistency. For real-world images without clean references, we use no-reference metrics, including Entropy~\cite{chen2019no}, CLIPIQA~\cite{wang2023exploring}, MUSIQ~\cite{ke2021musiq}, and BRISQUE~\cite{mittal2012no}. Since no-reference metrics may have different preferences across weather types and contents, qualitative comparisons are also provided for comprehensive evaluation.

\subsubsection{Implementation Details.}
All experiments are implemented in PyTorch 2.2.2~\cite{paszke2019pytorch} and conducted on a single NVIDIA GeForce RTX 4080 SUPER GPU. CFANet is trained with the AdamW optimizer~\cite{loshchilov2019iclr-decoupled} for 800K iterations. We adopt a progressive training strategy, where the input crop size is increased from $128\times128$ to $320\times320$, the batch size is reduced from 16 to 2, and the learning rate is decayed from $3\times10^{-4}$ to $1\times10^{-7}$.

\begin{table*}[t]
\centering
\caption{Quantitative comparison on Rain100H and Rain100L~\cite{yang2017deep}. Higher PSNR and SSIM and lower LPIPS and FID indicate better performance. Best and second-best results are shown in bold and underlined, respectively.}
\label{sys_rain}

\fontsize{8.5pt}{12pt}\selectfont
\setlength{\tabcolsep}{0.8pt}
\renewcommand{\arraystretch}{0.85}

\begin{tabular*}{\textwidth}{
    @{\extracolsep{\fill}}lcccccccc@{}
}
\toprule
& \multicolumn{4}{c}{\textit{Rain100H}}
& \multicolumn{4}{c}{\textit{Rain100L}} \\
\cmidrule(lr){2-5}
\cmidrule(lr){6-9}

Method
& PSNR$\uparrow$
& SSIM$\uparrow$
& LPIPS$\downarrow$
& FID$\downarrow$
& PSNR$\uparrow$
& SSIM$\uparrow$
& LPIPS$\downarrow$
& FID$\downarrow$ \\
\midrule

WeatherDiff~\cite{ozdenizci2023restoring}
& 25.50 & 0.823 & 0.220 & 96.754
& 27.46 & 0.873 & 0.220 & 93.377 \\

PromptIR~\cite{potlapalli2023promptir}
& 26.45 & 0.833 & 0.190 & 82.702
& 28.47 & 0.884 & 0.202 & 85.731 \\

DA-CLIP~\cite{luo2024controlling}
& 23.09 & 0.729 & 0.348 & 151.743
& 29.29 & 0.913 & 0.147 & 62.172 \\

DiffUIR~\cite{zheng2024selective}
& 26.25 & 0.814 & 0.223 & 93.097
& 31.12 & 0.933 & 0.114 & 44.026 \\

CycleRDM~\cite{xue2025unified}
& 25.54 & 0.798 & 0.246 & 104.333
& 31.63 & 0.937 & 0.101 & 38.943 \\

AWRaCLe~\cite{rajagopalan2025awracle}
& 26.69 & \underline{0.847}
& \underline{0.178} & \underline{76.088}
& 33.96 & 0.958 & 0.067 & 27.209 \\

AdaIR~\cite{cui2025adair}
& 26.64 & 0.831 & 0.199 & 79.274
& 34.41 & 0.961 & 0.067 & 24.955 \\

StarIR~\cite{cui2026starir}
& \underline{26.83} & 0.841
& 0.189 & 80.033
& \underline{35.68} & \underline{0.967}
& \underline{0.051} & \underline{20.070} \\

WeatherRemover~\cite{qu2025weatherremover}
& 25.17 & 0.829 & 0.201 & 88.248
& 31.97 & 0.947 & 0.098 & 37.410 \\

\midrule

CFANet (Ours)
& \textbf{27.36} & \textbf{0.858}
& \textbf{0.165} & \textbf{71.473}
& \textbf{35.98} & \textbf{0.972}
& \textbf{0.025} & \textbf{15.228} \\

\bottomrule
\end{tabular*}
\end{table*}

\begin{table}[t]
\centering
\caption{Quantitative comparison on SOTS-Outdoor~\cite{li2018benchmarking}. Higher PSNR and SSIM and lower LPIPS and FID indicate better performance. Best and
second-best results are shown in bold and underlined, respectively.}
\label{sys_sots}

\fontsize{10pt}{12pt}\selectfont
\setlength{\tabcolsep}{3pt}
\renewcommand{\arraystretch}{1.0}

\begin{tabular*}{\linewidth}{
    @{\extracolsep{\fill}}lcccc@{}
}
\toprule
Method
& PSNR$\uparrow$
& SSIM$\uparrow$
& LPIPS$\downarrow$
& FID$\downarrow$ \\
\midrule

WeatherDiff~\cite{ozdenizci2023restoring}
& 28.15 & 0.972 & 0.019 & 6.375 \\

PromptIR~\cite{potlapalli2023promptir}
& 30.36 & 0.982 & 0.014 & 4.119 \\

DA-CLIP~\cite{luo2024controlling}
& 27.88 & 0.969 & 0.022 & 6.101 \\

DiffUIR~\cite{zheng2024selective}
& \underline{30.95} & 0.946
& 0.040 & 10.518 \\

CycleRDM~\cite{xue2025unified}
& 29.09 & 0.974 & 0.017 & 5.038 \\

AWRaCLe~\cite{rajagopalan2025awracle}
& 30.46 & \underline{0.983}
& \underline{0.012} & 3.750 \\

AdaIR~\cite{cui2025adair}
& 30.25 & 0.981
& 0.013 & \underline{3.712} \\

StarIR~\cite{cui2026starir}
& 28.52 & 0.937 & 0.038 & 8.955 \\

WeatherRemover~\cite{qu2025weatherremover}
& 30.00 & 0.966 & 0.023 & 6.085 \\

\midrule
CFANet (Ours)
& \textbf{31.25} & \textbf{0.984}
& \textbf{0.012} & \textbf{3.228} \\
\bottomrule
\end{tabular*}
\end{table}

\begin{table*}[t]
\centering
\caption{Quantitative comparison on Snow100K-L, Snow100K-M, and Snow100K-S~\cite{liu2018desnownet}. Average denotes the arithmetic mean across the three subsets. Higher PSNR and SSIM and lower LPIPS and FID indicate better performance. Best and second-best results are shown in bold and underlined, respectively.}
\label{sys_snow}

\fontsize{9pt}{12pt}\selectfont
\setlength{\tabcolsep}{0.8pt}
\renewcommand{\arraystretch}{0.85}

\begin{tabular*}{\textwidth}{
    @{\extracolsep{\fill}}lcccccccc@{}
}
\toprule
& \multicolumn{4}{c}{\textit{Snow100K-L}}
& \multicolumn{4}{c}{\textit{Snow100K-M}} \\
\cmidrule(lr){2-5}
\cmidrule(lr){6-9}

Method
& PSNR$\uparrow$
& SSIM$\uparrow$
& LPIPS$\downarrow$
& FID$\downarrow$
& PSNR$\uparrow$
& SSIM$\uparrow$
& LPIPS$\downarrow$
& FID$\downarrow$ \\
\midrule

WeatherDiff~\cite{ozdenizci2023restoring}
& 27.95 & 0.872 & 0.125 & 6.259
& 32.74 & 0.938 & 0.065 & 2.278 \\

PromptIR~\cite{potlapalli2023promptir}
& 28.94 & 0.890 & 0.115 & 5.697
& 31.73 & 0.927 & 0.076 & 2.801 \\

DA-CLIP~\cite{luo2024controlling}
& 26.49 & 0.846 & 0.165 & 9.838
& 29.90 & 0.908 & 0.106 & 4.632 \\

DiffUIR~\cite{zheng2024selective}
& 28.04 & 0.878 & 0.137 & 6.561
& 31.76 & 0.931 & 0.083 & 2.965 \\

CycleRDM~\cite{xue2025unified}
& 29.82 & 0.903 & 0.104 & 5.232
& 33.66 & 0.946 & 0.058 & 2.148 \\

AWRaCLe~\cite{rajagopalan2025awracle}
& 30.32 & 0.909 & 0.088 & 3.459
& 34.16 & 0.949 & 0.050 & 1.690 \\

StarIR~\cite{cui2026starir}
& \underline{30.81} & \underline{0.914}
& \underline{0.085} & 3.588
& \underline{34.54} & \underline{0.952}
& \underline{0.047} & 1.568 \\

WeatherRemover~\cite{qu2025weatherremover}
& 30.27 & 0.899 & 0.093 & \textbf{3.411}
& 33.76 & 0.941 & 0.053 & \underline{1.525} \\

\midrule

CFANet (Ours)
& \textbf{31.10} & \textbf{0.919}
& \textbf{0.078} & \underline{3.449}
& \textbf{35.16} & \textbf{0.959}
& \textbf{0.042} & \textbf{1.457} \\

\bottomrule
\end{tabular*}

\par\vspace{4pt}

\begin{tabular*}{\textwidth}{
    @{\extracolsep{\fill}}lcccccccc@{}
}
\toprule
& \multicolumn{4}{c}{\textit{Snow100K-S}}
& \multicolumn{4}{c}{\textit{Average}} \\
\cmidrule(lr){2-5}
\cmidrule(lr){6-9}

Method
& PSNR$\uparrow$
& SSIM$\uparrow$
& LPIPS$\downarrow$
& FID$\downarrow$
& PSNR$\uparrow$
& SSIM$\uparrow$
& LPIPS$\downarrow$
& FID$\downarrow$ \\
\midrule

WeatherDiff~\cite{ozdenizci2023restoring}
& 34.23 & 0.948 & 0.053 & 1.689
& 31.64 & 0.919 & 0.081 & 3.409 \\

PromptIR~\cite{potlapalli2023promptir}
& 33.46 & 0.940 & 0.061 & 1.912
& 31.38 & 0.919 & 0.084 & 3.470 \\

DA-CLIP~\cite{luo2024controlling}
& 31.49 & 0.922 & 0.088 & 3.193
& 29.29 & 0.892 & 0.120 & 5.888 \\

DiffUIR~\cite{zheng2024selective}
& 33.64 & 0.943 & 0.068 & 2.155
& 31.15 & 0.917 & 0.096 & 3.894 \\

CycleRDM~\cite{xue2025unified}
& 35.30 & 0.955 & 0.047 & 1.681
& 32.93 & 0.935 & 0.070 & 3.020 \\

AWRaCLe~\cite{rajagopalan2025awracle}
& 35.78 & 0.958 & 0.040 & 1.305
& 33.42 & 0.939 & 0.059 & 2.151 \\

StarIR~\cite{cui2026starir}
& \underline{36.10} & \underline{0.960}
& \underline{0.038} & 1.226
& \underline{33.82} & \underline{0.942}
& \underline{0.057} & 2.127 \\

WeatherRemover~\cite{qu2025weatherremover}
& 35.36 & 0.951 & 0.043 & \underline{1.188}
& 33.13 & 0.930 & 0.063 & \underline{2.041} \\

\midrule

CFANet (Ours)
& \textbf{36.49} & \textbf{0.965}
& \textbf{0.033} & \textbf{1.174}
& \textbf{34.25} & \textbf{0.948}
& \textbf{0.051} & \textbf{2.027} \\

\bottomrule
\end{tabular*}
\end{table*}

\begin{figure}[t]
\centering
\centerline{\includegraphics[width=\textwidth,height=0.7\textwidth]{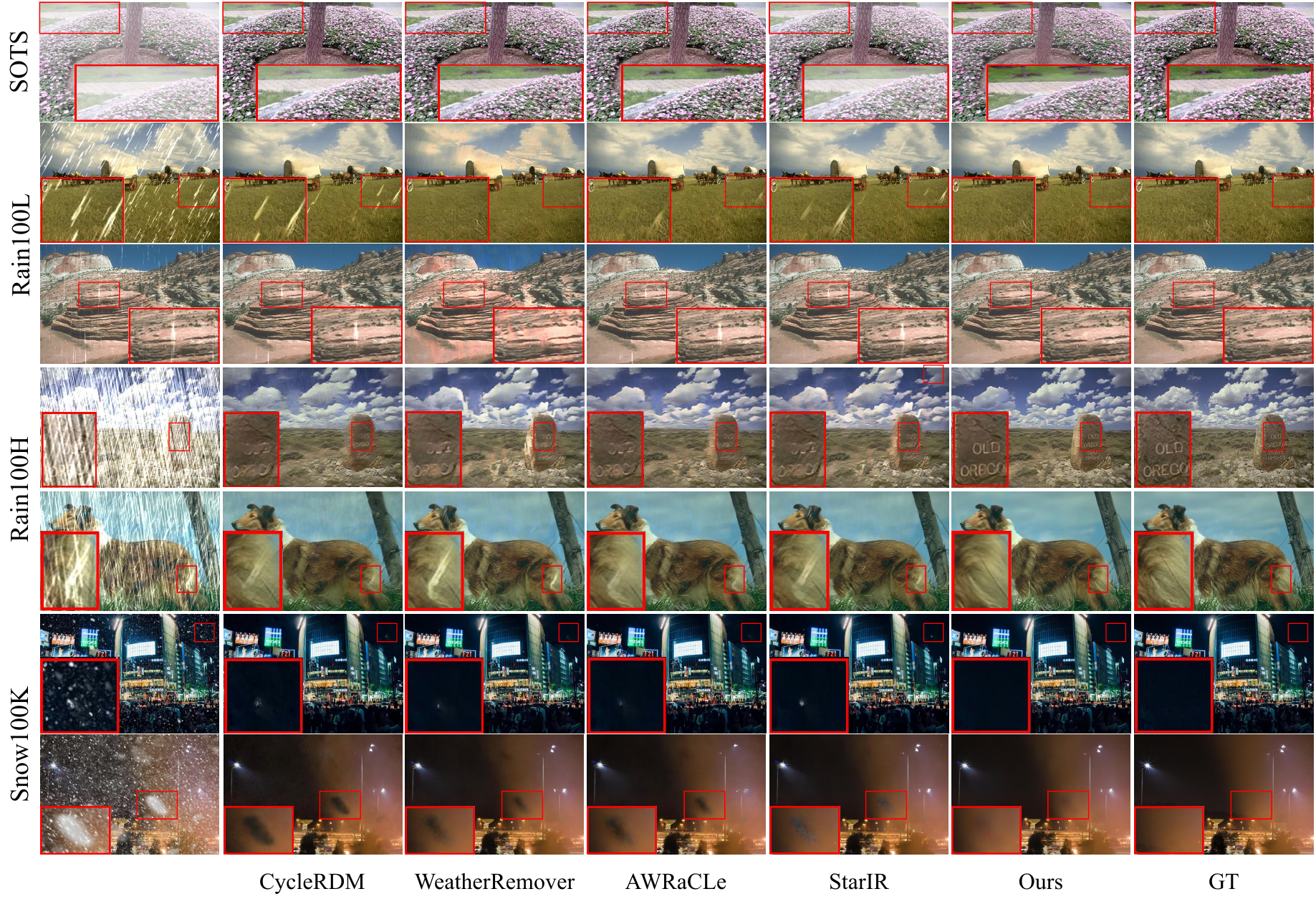}}
\caption{Qualitative comparison on synthetic single-weather benchmarks: SOTS-Outdoor~\cite{li2018benchmarking} for dehazing, Rain100L and Rain100H~\cite{yang2017deep} for deraining, and Snow100K~\cite{liu2018desnownet} for desnowing. Red boxes and enlarged views highlight local weather artifacts and scene details for comparison.}
\label{sys_pic}
\end{figure}

\subsection{Experiment Results}
The evaluation covers restoration under individual weather types, reconstruction under coexisting degradations, and restored appearance on real images. The paired benchmarks provide clean references for assessing reconstruction fidelity and perceptual similarity. The real-world evaluation combines no-reference measures with visual comparisons to examine outputs without clean targets.

\subsubsection{Single-Weather Restoration}
Tables~\ref{sys_rain}, \ref{sys_sots}, and \ref{sys_snow}  compares different methods on Rain100L, Rain100H~\cite{yang2017deep}, SOTS-Outdoor~\cite{li2018benchmarking}, and the three Snow100K subsets~\cite{liu2018desnownet}. Overall, CFANet achieves competitive reconstruction and perceptual performance across deraining, dehazing, and desnowing. In particular, compared with StarIR~\cite{cui2026starir} on Rain100L, CFANet improves PSNR by 0.30~dB and reduces LPIPS from 0.051 to 0.025, indicating that the reconstruction gain is accompanied by closer perceptual agreement with the clean reference. Similar improvements in PSNR, SSIM, and LPIPS over StarIR are observed across the Snow100K subsets. These results support the applicability of CFANet across the three evaluated weather tasks.

Fig.~\ref{sys_pic} presents visual comparisons on the synthetic single-weather benchmarks. Under dense rain or snow, some competing results retain visible streaks or particles, while fine structures become blurred in heavily degraded regions. CFANet suppresses these artifacts while preserving details such as the lettering on the sign and the building structures in the night scenes. On SOTS-Outdoor, it also recovers scene contrast and vegetation details obscured by haze.

\begin{table}[t]
\centering
\caption{Quantitative comparison on five weather categories of
CDD-11~\cite{guo2024onerestore}: Haze (H), Rain (R), Snow (S),
Haze+Rain (H+R), and Haze+Snow (H+S). Average denotes the
arithmetic mean over these five categories. Higher PSNR and
SSIM values indicate better performance. Best and second-best
results are shown in bold and underlined, respectively.}
\label{cdd-11}

\fontsize{10pt}{12pt}\selectfont
\setlength{\tabcolsep}{3pt}
\renewcommand{\arraystretch}{1.0}

\begin{tabular*}{\linewidth}{
    @{\extracolsep{\fill}}lcccccc@{}
}
\toprule
\multicolumn{7}{c}{\textit{Single-Weather Categories}} \\
\midrule
& \multicolumn{2}{c}{Haze (H)}
& \multicolumn{2}{c}{Rain (R)}
& \multicolumn{2}{c}{Snow (S)} \\
Method
& PSNR$\uparrow$ & SSIM$\uparrow$
& PSNR$\uparrow$ & SSIM$\uparrow$
& PSNR$\uparrow$ & SSIM$\uparrow$ \\
\midrule

PromptIR~\cite{potlapalli2023promptir}
& 26.10 & 0.969
& 31.56 & 0.946
& 31.53 & 0.960 \\

OneRestore~\cite{guo2024onerestore}
& 32.52 & 0.990
& 33.40 & 0.964
& 34.31 & 0.973 \\

OKNet~\cite{cui2024omni}
& 28.27 & 0.983
& 30.98 & 0.947
& 32.35 & 0.963 \\

BioIR~\cite{cui2026bio}
& 34.77 & 0.992
& 34.48 & 0.970
& 36.26 & 0.979 \\

CycleRDM~\cite{xue2025unified}
& 35.37 & 0.991
& 34.79 & 0.972
& 36.25 & 0.975 \\

UniRestore~\cite{chen2025unirestore}
& 35.29 & 0.992
& \textbf{37.29} & \underline{0.979}
& \underline{38.02} & \underline{0.985} \\

WeatherRemover~\cite{qu2025weatherremover}
& \underline{36.88} & \underline{0.995}
& 35.44 & 0.969
& 36.41 & 0.981 \\

\midrule

CFANet (Ours)
& \textbf{37.12} & \textbf{0.996}
& \underline{36.97} & \textbf{0.980}
& \textbf{38.34} & \textbf{0.989} \\

\bottomrule
\end{tabular*}

\par\vspace{4pt}

\begin{tabular*}{\linewidth}{
    @{\extracolsep{\fill}}lcccccc@{}
}
\toprule
\multicolumn{7}{c}{\textit{Compound-Weather Categories and Average}} \\
\midrule
& \multicolumn{2}{c}{H+R}
& \multicolumn{2}{c}{H+S}
& \multicolumn{2}{c}{Average} \\
Method
& PSNR$\uparrow$ & SSIM$\uparrow$
& PSNR$\uparrow$ & SSIM$\uparrow$
& PSNR$\uparrow$ & SSIM$\uparrow$ \\
\midrule

PromptIR~\cite{potlapalli2023promptir}
& 24.54 & 0.924
& 23.70 & 0.925
& 27.49 & 0.945 \\

OneRestore~\cite{guo2024onerestore}
& 29.99 & 0.957
& 30.21 & 0.964
& 32.09 & 0.970 \\

OKNet~\cite{cui2024omni}
& 25.83 & 0.934
& 27.14 & 0.949
& 28.91 & 0.955 \\

BioIR~\cite{cui2026bio}
& 31.64 & 0.968
& 31.60 & 0.972
& 33.75 & 0.976 \\

CycleRDM~\cite{xue2025unified}
& 31.74 & \underline{0.969}
& 31.75 & 0.947
& 33.98 & 0.971 \\

UniRestore~\cite{chen2025unirestore}
& \underline{33.14} & 0.945
& 31.85 & 0.934
& \underline{35.12} & 0.967 \\

WeatherRemover~\cite{qu2025weatherremover}
& 32.66 & 0.965
& \underline{32.72} & \underline{0.975}
& 34.82 & \underline{0.977} \\

\midrule

CFANet (Ours)
& \textbf{33.52} & \textbf{0.973}
& \textbf{33.24} & \textbf{0.978}
& \textbf{35.84} & \textbf{0.983} \\

\bottomrule
\end{tabular*}
\end{table}

\subsubsection{Compound-Weather Restoration}
Table~\ref{cdd-11} compares different methods on the five evaluated weather-related categories of CDD-11~\cite{guo2024onerestore}. CFANet achieves an average PSNR/SSIM of 35.84/0.983 over these categories. In particular, it improves PSNR over WeatherRemover by 0.86~dB on Haze+Rain and 0.52~dB on Haze+Snow. These gains support the applicability of the shared-private adaptation framework to the evaluated inputs containing coexisting weather degradations.

\begin{figure}[t]
\centering
\centerline{\includegraphics[width=\textwidth,height=0.58\textwidth]{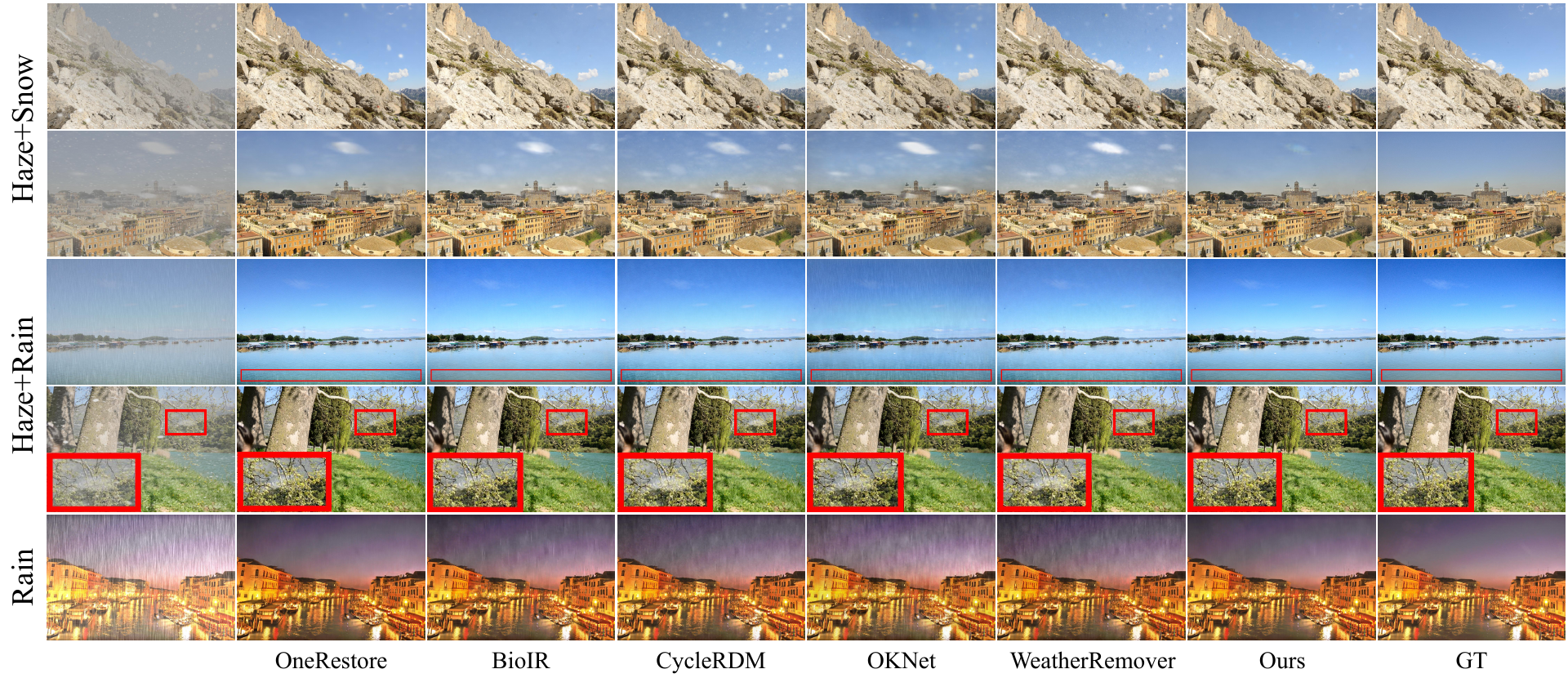}}
\caption{Qualitative comparison on CDD-11~\cite{guo2024onerestore}. Red boxes highlight local regions for comparison.}
\label{cdd_pic}
\end{figure}

Fig.~\ref{cdd_pic} presents qualitative results on CDD-11~\cite{guo2024onerestore}. Under Haze+Rain or Haze+Snow, both background visibility and local structures are affected. Although several compared methods recover much of the overall visibility, the restored outputs differ in local contrast and vegetation detail. CFANet uses non-exclusive weather routing to combine common and weather-private adapters, allowing responses to coexisting weather components to contribute jointly. In the selected compound examples, it recovers scene contrast while retaining distinct boundaries in the mountain and vegetation regions.

\begin{table}[t]
\centering
\caption{No-reference quantitative comparison on the real-world
dehazing benchmarks RTTS~\cite{li2018benchmarking} and
Fattal~\cite{fattal2014dehazing}. Higher Entropy, CLIPIQA,
and MUSIQ and lower BRISQUE values are preferred under the
reported evaluation criteria. Best and second-best results
are shown in bold and underlined, respectively.}
\label{real_hazy}

\fontsize{10pt}{12pt}\selectfont
\setlength{\tabcolsep}{3pt}
\renewcommand{\arraystretch}{1.0}

\begin{tabular*}{\linewidth}{
    @{\extracolsep{\fill}}llcccc@{}
}
\toprule
\multicolumn{6}{c}{\textit{RTTS~\cite{li2018benchmarking}}} \\
\midrule
Method & Venue\&Year
& Entropy$\uparrow$
& CLIPIQA$\uparrow$
& MUSIQ$\uparrow$
& BRISQUE$\downarrow$ \\
\midrule

GridFormer & IJCV'24
& \underline{7.129} & 0.420
& 56.751 & \underline{25.321} \\

DiffUIR & CVPR'24
& 7.056 & 0.423
& \underline{57.095} & 28.926 \\

CycleRDM & PR'25
& 7.061 & 0.431
& 56.896 & 27.953 \\

AdaIR & ICLR'25
& 7.080 & 0.418
& 56.050 & 25.533 \\

AWRaCLe & AAAI'25
& 7.094 & 0.419
& 56.273 & 27.160 \\

StarIR & TPAMI'26
& 7.086 & \underline{0.441}
& 56.343 & 29.229 \\

WeatherRemover & TAI'26
& 6.970 & 0.395
& 55.249 & 34.499 \\

\midrule

CFANet (Ours) & --
& \textbf{7.185} & \textbf{0.452}
& \textbf{57.219} & \textbf{24.936} \\

\bottomrule
\end{tabular*}

\par\vspace{4pt}

\begin{tabular*}{\linewidth}{
    @{\extracolsep{\fill}}llcccc@{}
}
\toprule
\multicolumn{6}{c}{\textit{Fattal~\cite{fattal2014dehazing}}} \\
\midrule
Method & Venue\&Year
& Entropy$\uparrow$
& CLIPIQA$\uparrow$
& MUSIQ$\uparrow$
& BRISQUE$\downarrow$ \\
\midrule

GridFormer & IJCV'24
& 7.182 & \underline{0.568}
& 63.400 & 17.441 \\

DiffUIR & CVPR'24
& 7.152 & 0.531
& 63.664 & 17.212 \\

CycleRDM & PR'25
& 7.183 & 0.525
& 60.257 & \underline{14.215} \\

AdaIR & ICLR'25
& 7.208 & 0.522
& 61.958 & 17.126 \\

AWRaCLe & AAAI'25
& \underline{7.268} & 0.507
& 63.187 & 14.531 \\

StarIR & TPAMI'26
& 7.046 & 0.513
& 60.404 & 24.556 \\

WeatherRemover & TAI'26
& 7.153 & 0.458
& \textbf{63.851} & 17.920 \\

\midrule

CFANet (Ours) & --
& \textbf{7.323} & \textbf{0.590}
& \underline{63.715} & \textbf{14.151} \\

\bottomrule
\end{tabular*}
\end{table}

\subsubsection{Real-World Restoration}
\begin{figure}[t]
\centering
\centerline{\includegraphics[width=\textwidth,height=0.35\textwidth]{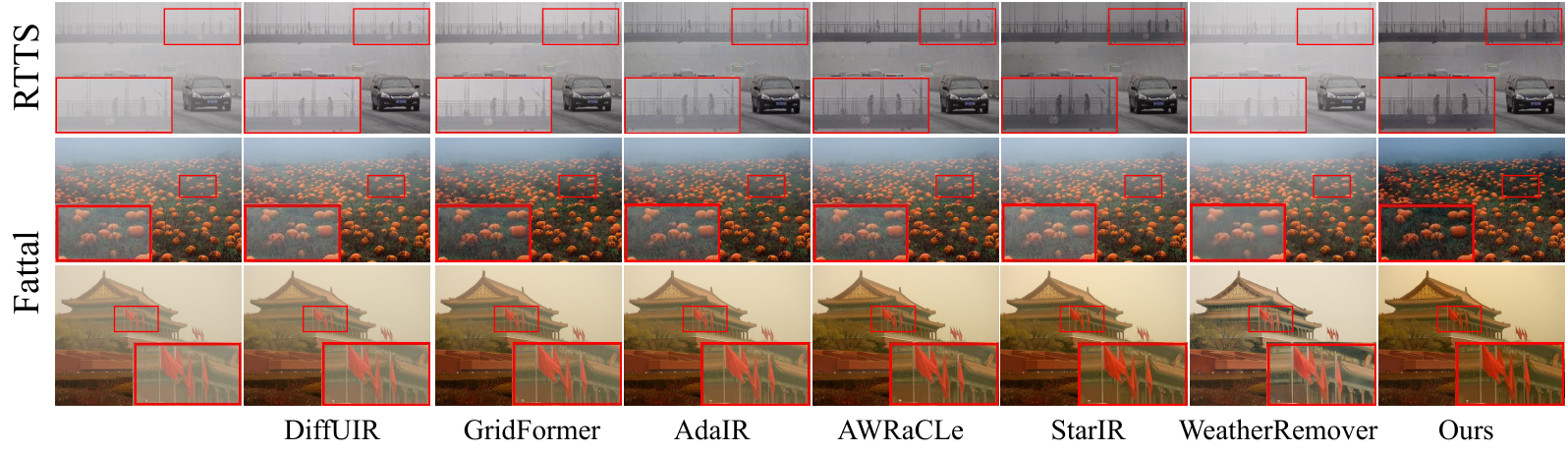}}
\caption{Qualitative comparison on the real-world dehazing benchmarks RTTS~\cite{li2018benchmarking} and Fattal~\cite{fattal2014dehazing}. Red boxes highlight local regions for comparison.}
\label{real_hazy_pic}
\end{figure}

\begin{figure}[t]
\centering
\centerline{\includegraphics[width=\textwidth,height=0.38\textwidth]{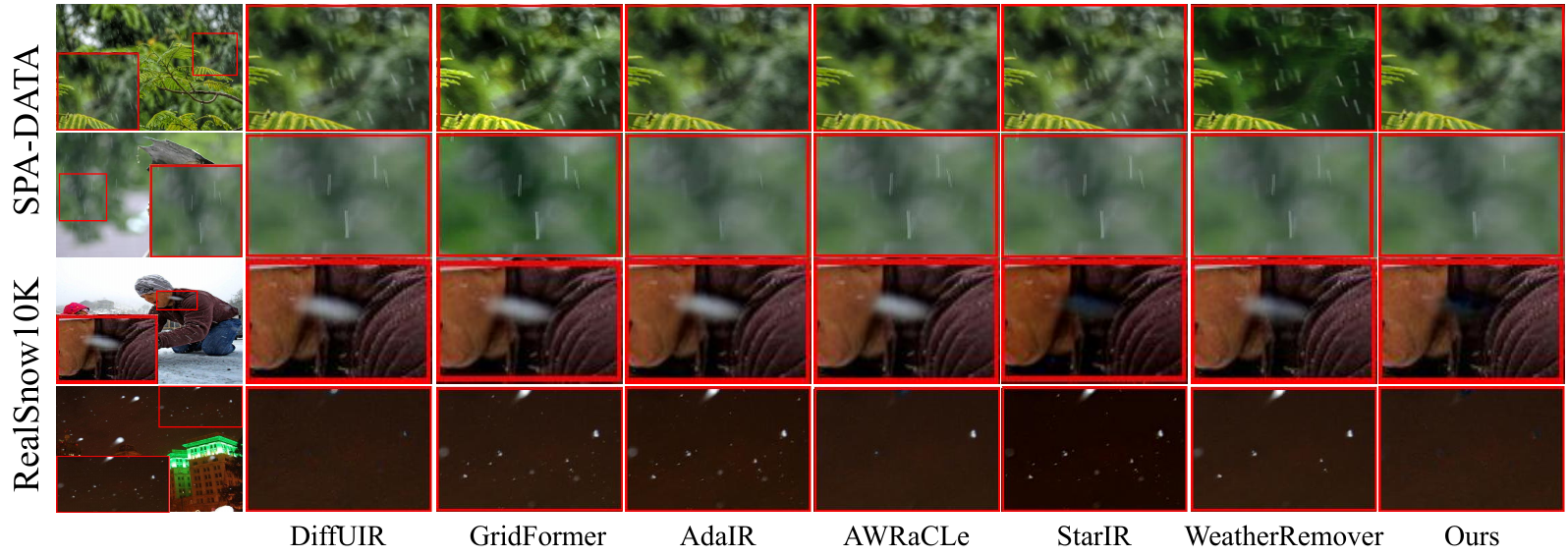}}
\caption{Qualitative comparison on SPA-DATA~\cite{wang2019spatial} for real-world deraining and RealSnow10K~\cite{lai2025snowmaster} for real-world desnowing. Red boxes highlight local regions for comparison.}
\label{real_rain_snow_pic}
\end{figure}

Tables~\ref{real_hazy} and~\ref{real_rain_snow} compare different methods on real-world dehazing, deraining, and desnowing benchmarks. Overall, CFANet achieves competitive no-reference quality, with consistent results across the four reported criteria on RTTS~\cite{li2018benchmarking} and RealSnow10K~\cite{lai2025snowmaster}.  On SPA-DATA~\cite{wang2019spatial}, CFANet obtains favorable Entropy and BRISQUE values, while its CLIPIQA/MUSIQ scores of 0.507/62.531 are below the 0.541/63.279 reported for GridFormer. These results indicate that the perceptual assessment of real-world restoration depends on both the weather condition and the quality criterion.
Figs.~\ref{real_hazy_pic} and~\ref{real_rain_snow_pic} present visual comparisons on real-world images. Dense haze reduces the visibility of background structures, while rain and snow introduce localized patterns around scene details. CFANet reduces visible haze and weather artifacts while retaining building outlines, foliage, and object boundaries.

\begin{table}[t]
\centering
\caption{No-reference quantitative comparison on
SPA-DATA~\cite{wang2019spatial} for real-world deraining
and RealSnow10K~\cite{lai2025snowmaster} for real-world
desnowing. Higher Entropy, CLIPIQA, and MUSIQ and lower
BRISQUE values are preferred under the reported evaluation
criteria. Best and second-best results are shown in bold
and underlined, respectively.}
\label{real_rain_snow}

\fontsize{10pt}{12pt}\selectfont
\setlength{\tabcolsep}{3pt}
\renewcommand{\arraystretch}{1.0}

\begin{tabular*}{\linewidth}{
    @{\extracolsep{\fill}}llcccc@{}
}
\toprule
\multicolumn{6}{c}{\textit{SPA-DATA~\cite{wang2019spatial}}} \\
\midrule
Method & Venue\&Year
& Entropy$\uparrow$
& CLIPIQA$\uparrow$
& MUSIQ$\uparrow$
& BRISQUE$\downarrow$ \\
\midrule

GridFormer & IJCV'24
& \underline{7.229} & \textbf{0.541}
& \textbf{63.279} & \underline{25.706} \\

DiffUIR & CVPR'24
& 7.082 & 0.434
& 59.010 & 28.360 \\

CycleRDM & PR'25
& 7.103 & 0.442
& 59.621 & 25.832 \\

AdaIR & ICLR'25
& 7.070 & 0.437
& 57.045 & 25.766 \\

AWRaCLe & AAAI'25
& 7.063 & 0.416
& 57.756 & 30.573 \\

StarIR & TPAMI'26
& 7.142 & 0.449
& 57.068 & 27.251 \\

WeatherRemover & TAI'26
& 7.108 & 0.431
& 58.290 & 30.932 \\

\midrule

CFANet (Ours) & --
& \textbf{7.255} & \underline{0.507}
& \underline{62.531} & \textbf{25.073} \\

\bottomrule
\end{tabular*}

\par\vspace{4pt}

\begin{tabular*}{\linewidth}{
    @{\extracolsep{\fill}}llcccc@{}
}
\toprule
\multicolumn{6}{c}{\textit{RealSnow10K~\cite{lai2025snowmaster}}} \\
\midrule
Method & Venue\&Year
& Entropy$\uparrow$
& CLIPIQA$\uparrow$
& MUSIQ$\uparrow$
& BRISQUE$\downarrow$ \\
\midrule

GridFormer & IJCV'2024
& 7.105 & 0.559
& 60.038 & 18.913 \\

DiffUIR & CVPR'2024
& 7.136 & 0.552
& 60.040 & 21.776 \\

CycleRDM & PR'2025
& 7.123 & 0.556
& \underline{61.324} & \underline{16.347} \\

AdaIR & ICLR'2025
& 7.171 & 0.574
& 60.317 & 18.674 \\

AWRaCLe & AAAI'2025
& 7.099 & 0.576
& 60.449 & 17.519 \\

StarIR & TPAMI'2026
& \underline{7.241} & \underline{0.595}
& 60.722 & 20.449 \\

WeatherRemover & TAI'2026
& 7.123 & 0.547
& 60.382 & 24.940 \\

\midrule

CFANet (Ours) & --
& \textbf{7.253} & \textbf{0.597}
& \textbf{61.432} & \textbf{16.054} \\

\bottomrule
\end{tabular*}
\end{table}

\subsubsection{Impact on Downstream Tasks}
To further investigate the practical utility of restored images, we evaluate CFANet on two downstream vision tasks, including object detection and semantic segmentation. Specifically, object detection is conducted on the RTTS dataset~\cite{li2018benchmarking} using a COCO-pretrained YOLOv8n detector~\cite{jocher2023ultralytics}, while semantic segmentation is evaluated on the Foggy Driving dataset~\cite{sakaridis2018semantic} using a Cityscapes-pretrained SegFormer-B0 model~\cite{xie2021segformer}. 

As shown in Table~\ref{downstream}, CFANet achieves the best performance in object detection, outperforming the unenhanced baseline by 4.65$\%$, 2.38$\%$, and 3.01$\%$ in precision, recall, and mAP$_{50}$, respectively. Compared with existing restoration methods, CFANet consistently delivers superior detection performance, achieving an mAP$_{50}$ of 53.50$\%$. For semantic segmentation, CFANet achieves the highest mIoU of 41.35$\%$ and the second-best mAcc of 54.09$\%$, improving upon the unenhanced baseline by 3.73 and 3.62 percentage points, respectively. These results suggest that CFANet not only effectively alleviates weather-induced degradations but also better preserves task-relevant visual structures and semantic information, thereby benefiting downstream visual understanding.

\begin{figure}[t]
\centering
\centerline{\includegraphics[width=\textwidth,height=0.48\textwidth]{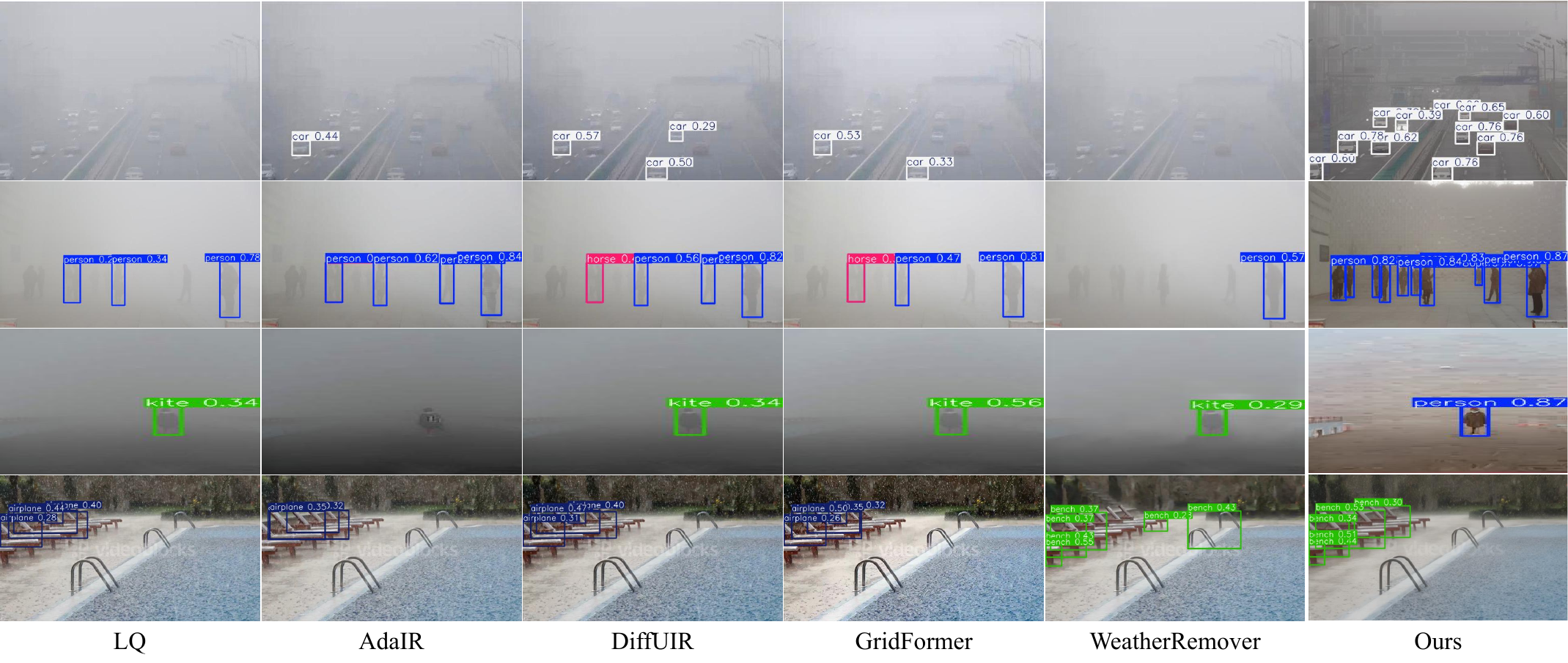}}
\caption{Qualitative object detection results on RTTS~\cite{li2018benchmarking} using YOLOv8n. Overlaid boxes, class labels, and confidence scores show the detector predictions.}
\label{jiance_pic}
\end{figure}

The qualitative detection results in Fig.~\ref{jiance_pic} further illustrate the impact of image restoration on downstream perception. Weather-induced degradations can obscure object boundaries and weaken discriminative visual cues, leading to missed detections or inaccurate object localization. By mitigating these degradations, CFANet provides clearer visual information for the detector, potentially facilitating more reliable object recognition and localization in challenging weather conditions.

\begin{table}[t]
\centering
\caption{Downstream object detection and semantic segmentation performance of different image restoration methods. P and R denote precision and recall, mAP$_{50}$ denotes mean average precision at an IoU threshold of 0.5. All scores are reported as percentages, with higher values indicating better performance. Best and second-best results for each metric are shown in bold and underlined, respectively.}
\normalsize
\setlength{\tabcolsep}{4pt}
\renewcommand{\arraystretch}{0.90}
\begin{tabular}{ccccccc}
\hline
                         &                               & \multicolumn{3}{c}{Detection}                                                              & \multicolumn{2}{c}{Segmentation}                            \\
\multirow{-2}{*}{Method} & \multirow{-2}{*}{Venue\&Year} & P$\uparrow$                  & R$\uparrow$                  & mAP50$\uparrow$              & mIoU$\uparrow$               & mAcc$\uparrow$               \\ \hline
w/o Enhancement          & -                             & 66.45                        & 45.13                        & 50.49                        & 37.62                        & 50.47                        \\
DiffUIR                  & CVPR'24                       & 66.84                        & 45.24                        & 50.78                        & 41.09                        & 53.49                        \\
GridFormer               & IJCV'24                       & 66.28                        & 46.19                        & 51.81                        & 41.27                        & \textbf{ 54.70} \\
AdaIR                    & ICLR'25                       & \underline{ 67.28} & \underline{ 47.15} & \underline{ 52.93} & \underline{ 41.30} & 54.02                        \\
WeatherRemover           & TAI'26                        & 66.09                        & 43.83                        & 49.14                        & 38.67                        & 52.05                        \\ \hline
CFANet (Ours)            & -                             & \textbf{ 71.10} & \textbf{ 47.51} & \textbf{ 53.50} & \textbf{ 41.35} & \underline{ 54.09} \\ \hline
\end{tabular}
\label{downstream}
\end{table}

\subsection{Ablation Study.}
Table~\ref{ablution_table} distinguishes the changes associated with the restoration block from those associated with weather-conditioned adaptation. CFABlocks without WMR and SPWA improve reconstruction across all three tasks relative to the baseline. Introducing WMR and SPWA together yields further gains, with the most pronounced change occurring in deraining, where PSNR increases from 29.18 to 30.55 dB. The comparison evaluates routing and private adaptation as a joint addition to the CFABlock backbone, making its contribution visible beyond the initial block replacement.

Output calibration and parameter regularization continue to affect reconstruction after routed adaptation is present. For deraining, NCA increases PSNR from 30.55 to 31.02 dB, and the final configuration labeled Full (+CFL) reaches 31.67 dB. Dehazing and desnowing show the same direction of change in PSNR and SSIM. These successive configurations intervene at different levels of the model: weather-dependent feature adaptation, global output calibration, and private-parameter regularization. The reported gains are retained when these operations are combined in the full model.

\begin{table}[t]
\centering
\normalsize
\caption{Component ablation of CFANet on dehazing, deraining, and desnowing. The configurations successively introduce CFABlocks, joint WMR and SPWA adaptation, NCA, and Conflict-Free Learning Strategy. The Full configuration includes all listed components. The dehazing results are evaluated on SOTS-Outdoor, the deraining results are averaged over Rain100H and Rain100L, and the desnowing results are averaged over the three Snow100K subsets.}
\setlength{\tabcolsep}{5pt}
\renewcommand{\arraystretch}{0.90}
\begin{tabular}{ccccccc}
\hline
                            &                          &                            & \multicolumn{2}{c}{+ CFABlock}                                                                                         &                         &                              \\ \cline{4-5}
\multirow{-2}{*}{Task}   & \multirow{-2}{*}{Metric} & \multirow{-2}{*}{Baseline} & \begin{tabular}[c]{@{}c@{}}w/o WMR \\ \& SPWA\end{tabular} & \begin{tabular}[c]{@{}c@{}}w/ WMR \\ \& SPWA\end{tabular} & \multirow{-2}{*}{+ NCA} & \multirow{-2}{*}{\begin{tabular}[c]{@{}c@{}}Full\\ (+CFL)\end{tabular}}        \\ \hline
                            & PSNR↑                    & 30.00                      & 30.34                                                      & 30.86                                                     & 31.04                   & \textbf{ 31.25} \\
\multirow{-2}{*}{Dehazing}  & SSIM↑                    & 0.966                      & 0.972                                                      & 0.978                                                     & 0.981                   & \textbf{ 0.984} \\ \hline
                            & PSNR↑                    & 28.57                      & 29.18                                                      & 30.55                                                     & 31.02                   & \textbf{ 31.67} \\
\multirow{-2}{*}{Deraining} & SSIM↑                    & 0.888                      & 0.895                                                      & 0.906                                                     & 0.910                   & \textbf{ 0.915} \\ \hline
                            & PSNR↑                    & 33.13                      & 33.39                                                      & 33.82                                                     & 34.05                   & \textbf{ 34.25} \\
\multirow{-2}{*}{Desnowing} & SSIM↑                    & 0.930                      & 0.935                                                      & 0.941                                                     & 0.945                   & \textbf{ 0.948} \\ \hline
\end{tabular}
\label{ablution_table}
\end{table}

\begin{table}[t]
\centering
\small
\caption{Model complexity comparison for $512\times512$ input images. Parameter counts and inference times are reported in millions (M) and milliseconds (ms), respectively.}
\setlength{\tabcolsep}{16pt}
\renewcommand{\arraystretch}{0.90}
\begin{tabular}{cccc}
\toprule
Method         & Venue\&Year & Params(M) & Runtimes(ms) \\ 
\midrule
WeatherDiff    & TPAMI'23    & 80.00     & 925.57       \\
DiffUIR        & CVPR'24     & 36.26     & 212.72       \\
AWRaCLe        & AAAI'25     & 186.46    & 84.67        \\
StarIR         & TPAMI'26    & 8.82      & 42.73        \\
WeatherRemover & TAI'26      & 24.32     & 65.30        \\
CFANet (Ours)  & -           & 28.01     & 98.53        \\ 
\bottomrule
\end{tabular}
\label{canshu}
\end{table}

\subsection{Model Complexity.}
Table~\ref{canshu} reports the complexity comparison on $512\times512$ inputs. CFANet has 28.01M parameters and requires 98.53 ms inference time on a single NVIDIA GeForce RTX 4080 SUPER GPU. Although CFANet introduces additional WMR, SPWA, and NCA modules, it remains much faster than diffusion-based DiffUIR~\cite{zheng2024selective} and substantially more compact than AWRaCLe~\cite{rajagopalan2025awracle}, showing a reasonable trade-off between restoration performance, compositional flexibility, and computational cost.

\subsection{Mechanism Analysis}
We further analyze two key mechanisms of CFANet: non-exclusive weather routing and conflict-free adapter disentanglement. 
Specifically, we visualize the weather composition weights predicted by WMR to verify its ability to activate multiple weather primitives for compound degradations. 
We also measure the similarity among weather-private adapters to examine whether the conflict-free learning objective encourages more distinguishable degradation-specific restoration directions.

\begin{figure}[t]
\centering
\centerline{\includegraphics[width=\textwidth,height=0.8\textwidth]{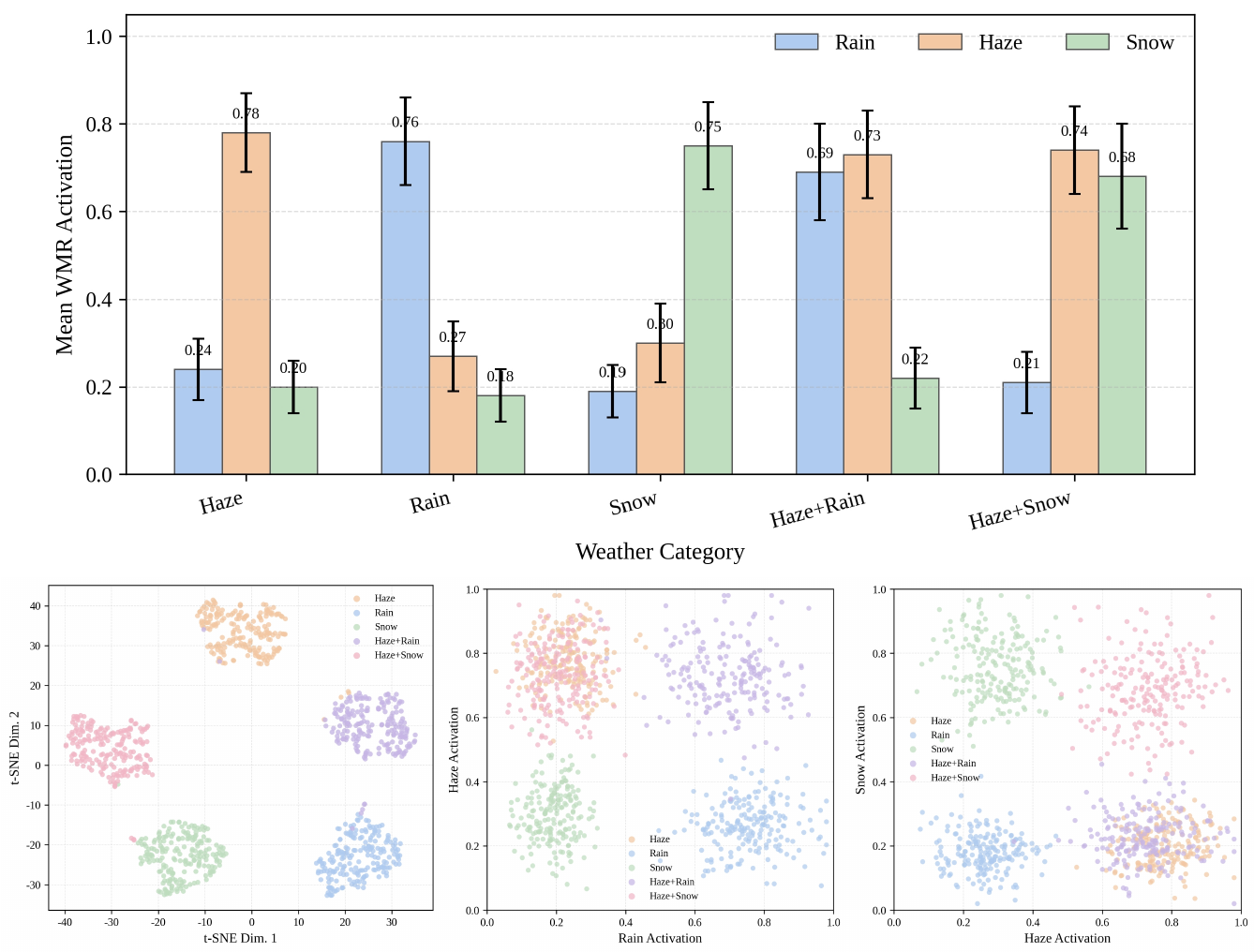}}
\caption{Analysis of the Weather Mixture Router (WMR). The top bar chart shows the mean predicted activation weights for rain, haze, and snow under different weather categories. The bottom visualizations show the t-SNE distribution and activation-space scatter plots, further demonstrating that WMR learns a structured and non-exclusive weather composition representation.}
\label{WMR_fenxi}
\end{figure}

\subsubsection{Analysis of Weather Mixture Router}
To further verify whether the Weather Mixture Router (WMR) learns meaningful composition-aware weather representations, we visualize the predicted weather activation weights in Fig.~\ref{WMR_fenxi}. 
For single-weather inputs, WMR assigns the highest activation to the corresponding weather primitive, such as haze, rain, or snow. 
For compound-weather inputs, including haze+rain and haze+snow, WMR activates multiple relevant weather primitives simultaneously instead of selecting only one dominant category. 
This confirms that WMR behaves as a non-exclusive composition router and supports the compositional restoration ability of CFANet.

The t-SNE visualization further shows that samples from different weather categories form distinguishable clusters, while compound-weather samples are located in regions related to their constituent weather primitives. 
The activation-space scatter plots also demonstrate that compound-weather samples receive high responses along multiple weather dimensions. 
These observations indicate that WMR learns a structured weather composition space rather than a single-label weather classification space.

\begin{figure}[t]
\centering
\centerline{\includegraphics[width=\textwidth,height=0.58\textwidth]{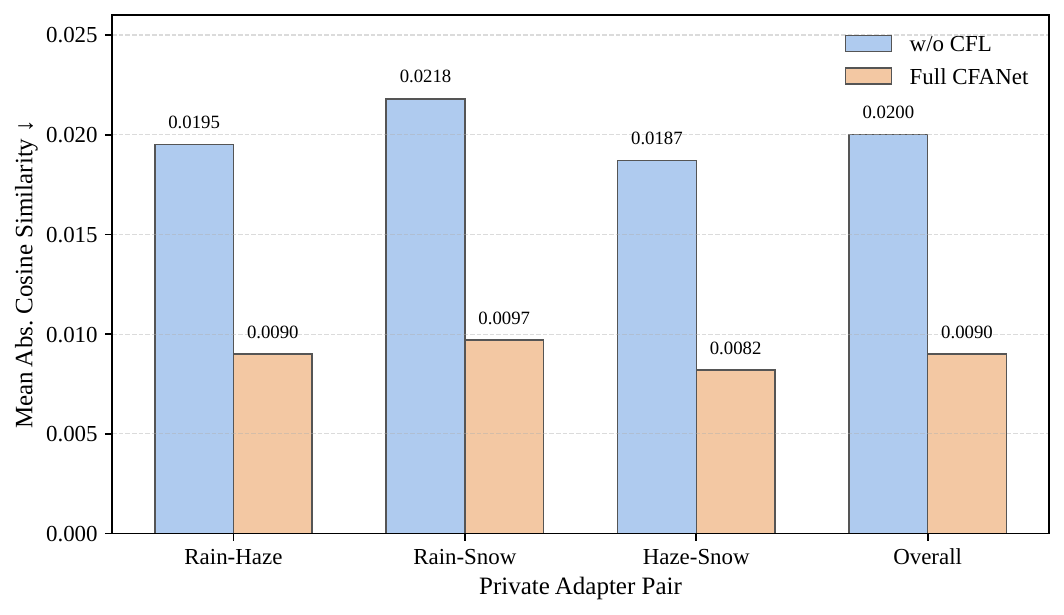}}
\caption{Analysis of private adapter disentanglement. 
We report the mean absolute cosine similarity between different weather-private adapter pairs. Compared with the variant without conflict-free learning (w/o CFL), the full CFANet substantially reduces the similarity among rain-, haze-, and snow-private adapters.
}
\label{private_adapter_similarity}
\end{figure}

\subsubsection{Analysis of Private Adapter Disentanglement}
We further analyze whether the proposed conflict-free learning objective encourages different weather-private adapters to learn distinguishable restoration directions. 
Specifically, we compute the mean absolute cosine similarity between the parameters of different private adapter pairs, including rain--haze, rain--snow, and haze--snow. 
As shown in Fig.~\ref{private_adapter_similarity}, the full CFANet consistently achieves lower similarity than the variant without conflict-free learning, indicating that the private adapters become less redundant and more disentangled.

Lower adapter similarity suggests that different weather-private adapters capture more distinct degradation-specific transformations. The orthogonality regularization encourages more distinguishable weather-specific adapter parameters, thereby helping mitigate interference among heterogeneous restoration tasks.

\section{Conclusion and Further Work}
We proposed CFANet for open-compositional all-weather image restoration. CFANet models adverse weather as a flexible composition of weather primitives through non-exclusive weather mixture routing, and dynamically combines common and weather-private adapters for composition-aware restoration. A lightweight Natural Color Adapter is further introduced to improve color fidelity and visual naturalness. Experiments on individual-weather and compound-weather benchmarks demonstrate competitive reconstruction fidelity and perceptual quality, while evaluations on real-world images and downstream vision tasks suggest that the restored results can benefit visual perception. The component ablations and adapter-similarity analysis further support the effectiveness of composition-aware routing and weather-specific parameter separation. These results show that shared-private adapter composition is a promising direction for unified restoration under diverse and mixed weather degradations. 

Despite these strengths, several limitations remain. The current router is restricted to three predefined weather primitives, and its generalization to unseen degradation types or more complicated compositions has not been established. Future work will investigate spatially adaptive and expandable composition routing, extend the primitive set to low-light and blur, and evaluate more challenging triple-weather combinations. We also aim to improve the efficiency of adapter composition and explore temporal consistency in video restoration. By explicitly separating common and weather-specific representations, CFANet provides a practical foundation for studying compositional adaptation in unified image restoration.

\bibliographystyle{elsarticle-num}
\bibliography{CFANet}



\end{document}